\documentclass[10pt,twocolumn,letterpaper]{article}
\pdfoutput=1

\usepackage[pagenumbers]{cvpr}
\makeatletter
\@namedef{ver@everyshi.sty}{}
\makeatother

\usepackage{graphicx}
\usepackage{amsmath}
\usepackage{amssymb}
\usepackage{booktabs}
\usepackage{algorithm} 
\usepackage{algpseudocode}
\usepackage{multirow}
\usepackage{array}
\usepackage{mathtools}
\usepackage{dcolumn}
\usepackage{nccmath}
\usepackage[export]{adjustbox}
\usepackage{graphbox}
\usepackage{caption}
\usepackage{enumitem}

\usepackage[pagebackref,breaklinks,colorlinks]{hyperref}
\usepackage[accsupp]{axessibility}  % Improves PDF readability for those with disabilities.

\usepackage[capitalize]{cleveref}
\crefname{section}{Sec.}{Secs.}
\crefname{table}{Tab.}{Tabs.}
\crefname{equation}{Eq.}{Eqs.}
\crefname{figure}{Fig.}{Figs.}
\crefname{algorithm}{Alg.}{Algs.}
\Crefname{section}{Sec.}{Secs.}
\Crefname{table}{Tab.}{Tabs.}
\Crefname{equation}{Eq.}{Eqs.}
\Crefname{figure}{Fig.}{Figs.}
\Crefname{algorithm}{Alg.}{Algs.}

\begin{document}


\title{Ham2Pose: Animating Sign Language Notation into Pose Sequences}

\author{Rotem Shalev Arkushin\\
Reichman University\\
{\tt\small rotemroo@gmail.com}
\and
Amit Moryossef\\
Bar-Ilan University\\
{\tt\small amitmoryossef@gmail.com}
\and
Ohad Fried\\
Reichman University\\
{\tt\small ofried@runi.ac.il}
\and
\small\url{https://rotem-shalev.github.io/ham-to-pose}
}

\maketitle

\newcommand{\betweencellpdf}{\ensuremath{h^c}}
\newcommand{\betweencellpdffine}{\ensuremath{h^f}}
\newcommand{\approxbetweencellpdffine}{\ensuremath{\hat{h}^f}}
\newcommand{\incellpdf}{\ensuremath{f}}
\newcommand{\incellpdfnormalized}{\ensuremath{f'}}

\newcommand{\incellcdf}{\ensuremath{F}}

\newcommand{\totalpdf}{\ensuremath{f_{dd}}}
\newcommand{\totalcdf}{\ensuremath{F_{dd}}}

\newcommand{\ignorethis}[1]{}
\newcommand{\redund}[1]{#1}

\newcommand{\apriori    }     {\textit{a~priori}}
\newcommand{\aposteriori}     {\textit{a~posteriori}}
\newcommand{\perse      }     {\textit{per~se}}
\newcommand{\naive      }     {{na\"{\i}ve}}
\newcommand{\Naive      }     {{Na\"{\i}ve}}
\newcommand{\Identity   }     {\mat{I}}
\newcommand{\Zero       }     {\mathbf{0}}
\newcommand{\Reals      }     {{\textrm{I\kern-0.18em R}}}
\newcommand{\isdefined  }     {\mbox{\hspace{0.5ex}:=\hspace{0.5ex}}}
\newcommand{\texthalf   }     {\ensuremath{\textstyle\frac{1}{2}}}
\newcommand{\half       }     {\ensuremath{\frac{1}{2}}}
\newcommand{\third      }     {\ensuremath{\frac{1}{3}}}
\newcommand{\fourth     }     {\ensuremath{\frac{1}{4}}}

\newcommand{\Lone} {\ensuremath{L_1}}
\newcommand{\Ltwo} {\ensuremath{L_2}}

\newcommand{\degree} {\ensuremath{^{\circ}}}

\newcommand{\mat        } [1] {{\text{\boldmath $\mathbit{#1}$}}}
\newcommand{\Approx     } [1] {\widetilde{#1}}
\newcommand{\change     } [1] {\mbox{{\footnotesize $\Delta$} \kern-3pt}#1}

\newcommand{\Order      } [1] {O(#1)}
\newcommand{\set        } [1] {{\lbrace #1 \rbrace}}
\newcommand{\floor      } [1] {{\lfloor #1 \rfloor}}
\newcommand{\ceil       } [1] {{\lceil  #1 \rceil }}
\newcommand{\inverse    } [1] {{#1}^{-1}}
\newcommand{\transpose  } [1] {{#1}^\mathrm{T}}
\newcommand{\invtransp  } [1] {{#1}^{-\mathrm{T}}}
\newcommand{\relu       } [1] {{\lbrack #1 \rbrack_+}}

\newcommand{\abs        } [1] {{| #1 |}}
\newcommand{\Abs        } [1] {{\left| #1 \right|}}
\newcommand{\norm       } [1] {{\| #1 \|}}
\newcommand{\Norm       } [1] {{\left\| #1 \right\|}}
\newcommand{\pnorm      } [2] {\norm{#1}_{#2}}
\newcommand{\Pnorm      } [2] {\Norm{#1}_{#2}}
\newcommand{\inner      } [2] {{\langle {#1} \, | \, {#2} \rangle}}
\newcommand{\Inner      } [2] {{\left\langle \begin{array}{@{}c|c@{}}
                               \displaystyle {#1} & \displaystyle {#2}
                               \end{array} \right\rangle}}

\newcommand{\twopartdef}[4]
{
  \left\{
  \begin{array}{ll}
    #1 & \mbox{if } #2 \\
    #3 & \mbox{if } #4
  \end{array}
  \right.
}

\newcommand{\fourpartdef}[8]
{
  \left\{
  \begin{array}{ll}
    #1 & \mbox{if } #2 \\
    #3 & \mbox{if } #4 \\
    #5 & \mbox{if } #6 \\
    #7 & \mbox{if } #8
  \end{array}
  \right.
}

\newcommand{\len}[1]{\text{len}(#1)}

\newlength{\w}
\newlength{\h}
\newlength{\x}

\definecolor{darkred}{rgb}{0.7,0.1,0.1}
\definecolor{darkgreen}{rgb}{0.1,0.6,0.1}
\definecolor{cyan}{rgb}{0.7,0.0,0.7}
\definecolor{otherblue}{rgb}{0.1,0.4,0.8}
\definecolor{maroon}{rgb}{0.76,.13,.28}
\definecolor{burntorange}{rgb}{0.81,.33,0}

\ifdefined\ShowNotes
  \newcommand{\colornote}[3]{{\color{#1}\textbf{#2} #3\normalfont}}
\else
  \newcommand{\colornote}[3]{}
\fi

\newcommand {\todo}[1]{\colornote{cyan}{TODO}{#1}}
\newcommand {\ohad}[1]{\colornote{burntorange}{OF:}{#1}}
\newcommand {\rotem}[1]{\colornote{darkgreen}{RS:}{#1}}
\newcommand {\amit}[1]{\colornote{otherblue}{AM:}{#1}}

\newcommand {\reqs}[1]{\colornote{red}{\tiny #1}}

\newcommand {\new}[1]{\colornote{red}{#1}}

\newcommand*\rot[1]{\rotatebox{90}{#1}}

\newcommand {\newstuff}[1]{#1}

\newcommand\todosilent[1]{}

\newcommand{\woBGmask}{{w/o~bg~\&~mask}}
\newcommand{\woMask}{{w/o~mask}}

% Keywords command
\providecommand{\keywords}[1]
{
  %\small	
  \textbf{\textit{Keywords---}} #1
}

\begin{abstract}

Translating spoken languages into Sign languages is necessary for open communication between the hearing and hearing-impaired communities. To achieve this goal, we propose the first method for animating a text written in HamNoSys, a lexical Sign language notation, into signed pose sequences. 
As HamNoSys is universal by design, our proposed method offers a generic solution invariant to the target Sign language. 
Our method gradually generates pose predictions using transformer encoders that create meaningful representations of the text and poses while considering their spatial and temporal information. We use weak supervision for the training process and show that our method succeeds in learning from partial and inaccurate data.
Additionally, we offer a new distance measurement that considers missing keypoints, to measure the distance between pose sequences using \emph{DTW-MJE}. 
We validate its correctness using AUTSL, a large-scale Sign language dataset, show that it measures the distance between pose sequences more accurately than existing measurements, and use it to assess the quality of our generated pose sequences.
Code for the data pre-processing, the model, and the distance measurement is publicly released for future research.
 
\end{abstract}
\section{Introduction}
\label{sec:intro}

Sign languages are an important communicative tool within the deaf and hard-of-hearing (DHH) community and a central property of Deaf culture. 
According to the World Health Organization, there are more than 70 million deaf people worldwide \cite{who2021}, who collectively use more than 300 different Sign languages \cite{un2022}.
Using the visual-gestural modality to convey meaning, Sign languages are considered natural languages \cite{sandler2006sign}, with their own grammar and lexicons.
They are not universal and are mostly independent of spoken languages. For example, American Sign Language (ASL)---used predominantly in the United States---and British Sign Language (BSL)---used predominantly in the United Kingdom---are entirely different, despite English being the predominant spoken language in both. 
As such, the translation task between each signed and spoken language pair is different and requires different data. Building a robust system that translates spoken languages into Sign languages and vice versa is fundamental to alleviate communication gaps between the hearing-impaired and the hearing communities. \begin{figure}[t]
    \centering
    \begin{tabular}{@{}c@{}c@{}c}
    \resizebox{0.47\linewidth}{!}{
        \begin{tabular}{ll}
            \textbf{Gloss}       & HOUSE3 \\[0.5mm]
            \textbf{HamNoSys}    & 
            \includegraphics[height=4mm, clip]{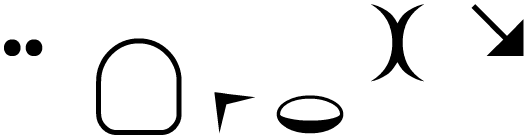}\\
            \textbf{SignWriting} & \includegraphics[height=7mm, clip, trim=.5mm .5mm .5mm .5mm]{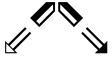}
        \end{tabular}
    } &
    \includegraphics[valign=c,width=.24\linewidth,frame]{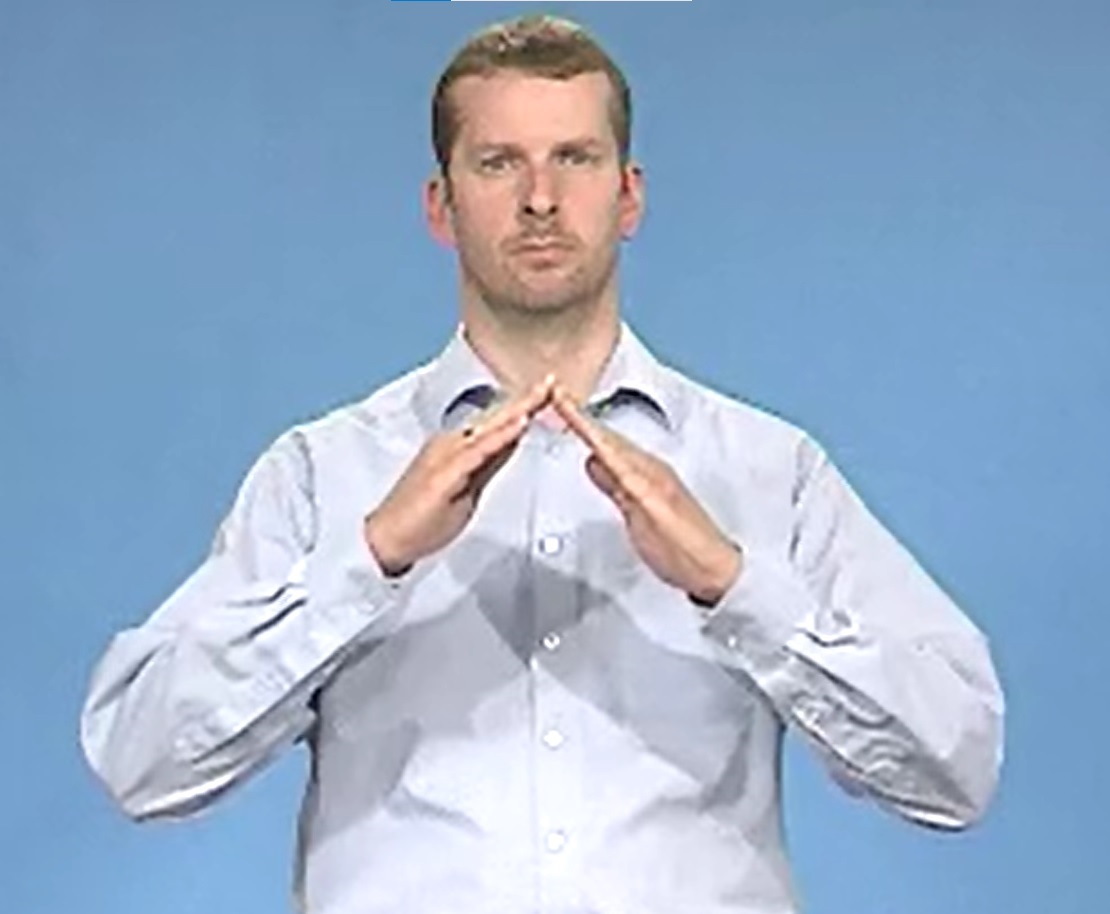} &
    \includegraphics[valign=c,width=.24\linewidth,frame]{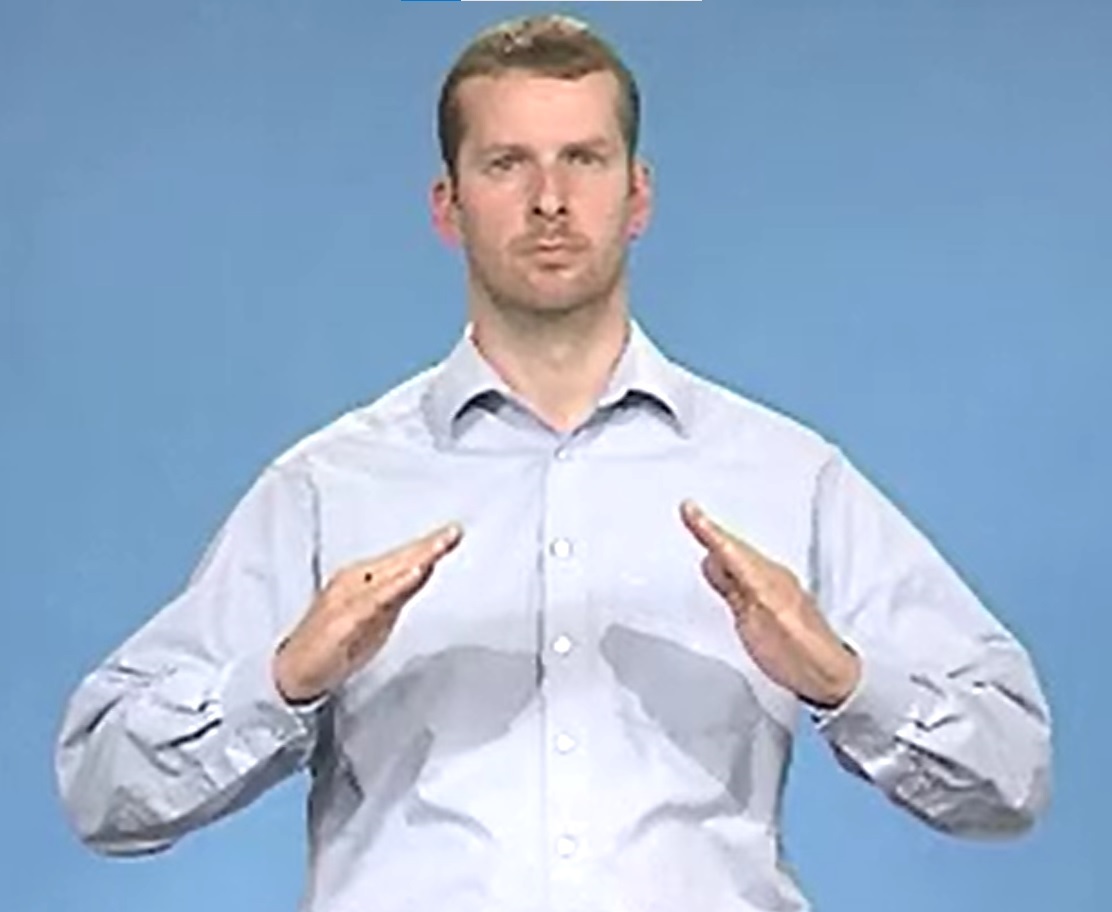}
    \end{tabular}
    
    \caption{
    \textbf{German Sign Language sign for ``Haus''.} \\
    \textbf{Gloss} is a unique semantic identifier;
    \textbf{HamNoSys} and \textbf{SignWriting} describe the phonology of a sign: \emph{Two flat hands with fingers closed, rotated towards each other, touching, then symmetrically moving diagonally downwards.}
    }
    \label{fig:notations}
\vspace{-0.45cm}
\end{figure}

% link to the original video:
%     \url{https://www.sign-lang.uni-hamburg.de/meinedgs/types/type14909\_en.html\#type63806.}}

% M547x578S15a11503x527S15a19478x527S22b05524x550S22b13455x550
While translation research from Sign languages into spoken languages has rapidly advanced in recent years \cite{rastgoo2020hand,rastgoo2022real,camgoz2018neural,camgoz2020sign,parelli2020exploiting,albanie2020bsl}, translating spoken languages into Sign languages, also known as Sign Language Production (SLP), remains a challenge \cite{stoll2018sign,stoll2020text2sign,saunders2020progressive,saunders2021mixed}. 
This is partially due to a misconception that deaf people are comfortable reading spoken language and do not require translation into Sign language.
However, there is no guarantee that someone whose first language is, for example, BSL, exhibits high literacy in written English.
SLP is usually done through an intermediate notation system such as a semantic notation system, \eg gloss (\Cref{subsec:semantic_notations}), or a lexical notation system, \eg HamNoSys, SignWriting (\Cref{subsec:lexical_notations}). The spoken language text is translated into the intermediate notation, which is then translated into the relevant signs. The signs can either be animated avatars or pose sequences later converted into videos.
Previous work has shown progress in translating spoken language text to Sign language lexical notations, namely HamNoSys \cite{walsh2022changing} and SignWriting \cite{zifan2022signwriting}, and in converting pose sequences into videos \cite{chan2019everybody, wang2018high, saunders2022signing}. There has been some work on animating HamNoSys into avatars \cite{bangham2000virtual, zwitserlood2004synthetic, efthimiou2010dicta, ebling2016building}, with unsatisfactory results (\Cref{subsec:avatar}), 
but no work on the task of animating HamNoSys into pose sequences.
Hence, in this work, we focus on animating HamNoSys into signed pose sequences, thus facilitating the task of SLP with a generic solution for all Sign languages. To do this, we collect and combine data from multiple HamNoSys-to-video datasets \cite{prillwitz2008dgs, matthes2012dicta, linde2014corpus},
extract pose keypoints from the videos using a pose estimation model, and process these further as detailed in \Cref{subsec:preprocess}.
We use the pose features as weak labels to train a model that gets HamNoSys text and a single pose frame as inputs and gradually generates the desired pose sequence from them. Despite the pose features being inaccurate and incomplete, our model still learns to produce the correct motions.
Additionally, we offer a new distance measurement that considers missing keypoints, to measure the distance between pose sequences using \emph{DTW-MJE} \cite{huang2021towards}.
We validate its correctness using AUTSL, a large-scale Sign language dataset \cite{sincan2020autsl}, and show that it measures pose sequences distance more accurately than currently used measurements.
Overall, our main contributions are: \vspace{-0.25cm}
\begin{enumerate}[leftmargin=*]
    \item We propose the first method for animating HamNoSys into pose sequences. \vspace{-0.25cm}
    \item We offer a new pose sequences distance measurement, validated on a large annotated dataset. \vspace{-0.25cm}
    \item We combine existing datasets, converting them to one enhanced dataset with processed pose features.
\end{enumerate}

\section{Background}
\label{sec:bg}

It is common to use an intermediate notation system for the SLP task. We discuss three of such notations below and show an example of them in
\Cref{fig:notations}.

\subsection{Semantic Notation Systems}
\label{subsec:semantic_notations}
Semantic notation systems are the most popular form of Sign language annotation. They treat Sign languages as discrete and annotate meaning units. A well-known semantic notation system is Gloss, a notation of the meaning of a word. 
% It may be in the language of the text or the reader's language. 
In Sign languages, glosses are usually sign-to-word transcriptions, where every sign has a unique identifier written in a spoken language. 
% for example, American Sign Language glosses would usually be written in English. 
Some previous works focused on translating spoken language into glosses \cite{saunders2020progressive, stoll2018sign, stoll2020text2sign}. The glosses are then usually transformed into videos using a gloss-to-video dictionary.
However, since glosses are Sign language-specific, each translation task is different from the other and requires different data.

\subsection{Lexical Notation Systems}
\label{subsec:lexical_notations}
Lexical notation systems annotate phonemes of Sign languages and can be used to transcribe any sign. Given that all the details about how to produce a sign are in the transcription itself,
% without assigning meaning to it, 
these notations can be universal and can be used to transcribe every Sign language.
Two examples of such universal notations are Hamburg Sign Language Notation System (HamNoSys) \cite{hanke2004hamnosys}, and SignWriting \cite{sutton1974signwriting}. They 
% are universal transcription systems for all Sign languages, with 
have a direct correspondence between symbols (glyphs) and gesture aspects, such as hand location, orientation, shape, and movement.
As they are not language-specific, it allows a Sign language invariant transcription for any desired sign.
% Another lexical notation system is SignWriting \cite{sutton1974signwriting}. Similarly to HamNoSys, it is universal and composed of symbols for hand shapes, orientation, location, and movements. However, 
Unlike HamNoSys, SignWriting is written in a spatial arrangement 
% on the page 
that does not follow a sequential order. While this arrangement was designed for better human readability, it is less ``friendly'' for computers, expecting sequential text order.
As demonstrated in \Cref{fig:notations}, HamNoSys is very expressive, precise, and easy to learn and use. Moreover, each part of it (each glyph) is responsible for one aspect of the sign, similarly to parts-of-speech in a spoken language sentence. 
Therefore, each HamNoSys sequence can be thought of as a ``sentence” and each glyph as a dictionary ``word” from the corpus of HamNoSys glyphs.

\subsection{Sign Language Notation Data}
Many Sign language corpora are annotated with Glosses \cite{forster2012rwth, prillwitz2008dgs, sincan2020autsl, duarte2021how2sign, li2020word}. However, as there is no single standard for gloss annotation, each corpus has its own unique identifiers for each sign at different granularity levels. 
This lack of universality---both in annotation guidelines and in different language data---makes it difficult to use and combine multiple corpora and design impactful translation systems.
% While the SignWriting notation system was designed before HamNoSys, 
Furthermore, since SignWriting is used more to ``write'' Sign language on the page, and not for video transcription, existing SignWriting resources \cite{sutton2002signbank, sign2mint, ASLwiki} usually include parallel SignWriting and spoken language text, without including parallel videos of the performed signs.
HamNoSys, on the other hand, was designed more for annotating existing Sign language videos, and as such, resources including HamNoSys \cite{matthes2012dicta, prillwitz2008dgs, linde2014corpus} always include parallel videos.
Compared to resources including gloss annotations, these resources are small, with only hundreds to thousands of signs with high-quality annotation. However, the language universality and annotation cohesion in these corpora allow grouping them together for the usage of all data without any annotation modification.
\section{Related Work}
\label{sec:prev_work}

In this section, we review related work in the field of SLP. We cover avatar approaches using HamNoSys as input and gloss approaches. Moreover, we cover HamNoSys generation work, including translating spoken language text or videos into HamNoSys, as these tasks allow for a full translation pipeline together with our work.
Furthermore, we mention diffusion models as a source of inspiration for our method, and text-to-motion works, explaining how our problem and data are different
from them. 

\subsection{Avatar Approaches}
\label{subsec:avatar}

\begin{figure}[t]
  \centering
    \begin{tabular}{@{}cc|c|c@{}c}
    & H-in-H & H-in-C & 
    \multicolumn{2}{c}{Wrong signing}  \\
      \rotatebox[origin=c]{90}{original} &
      \includegraphics[clip,align=c,height=1.8cm,width=1.38cm,frame]{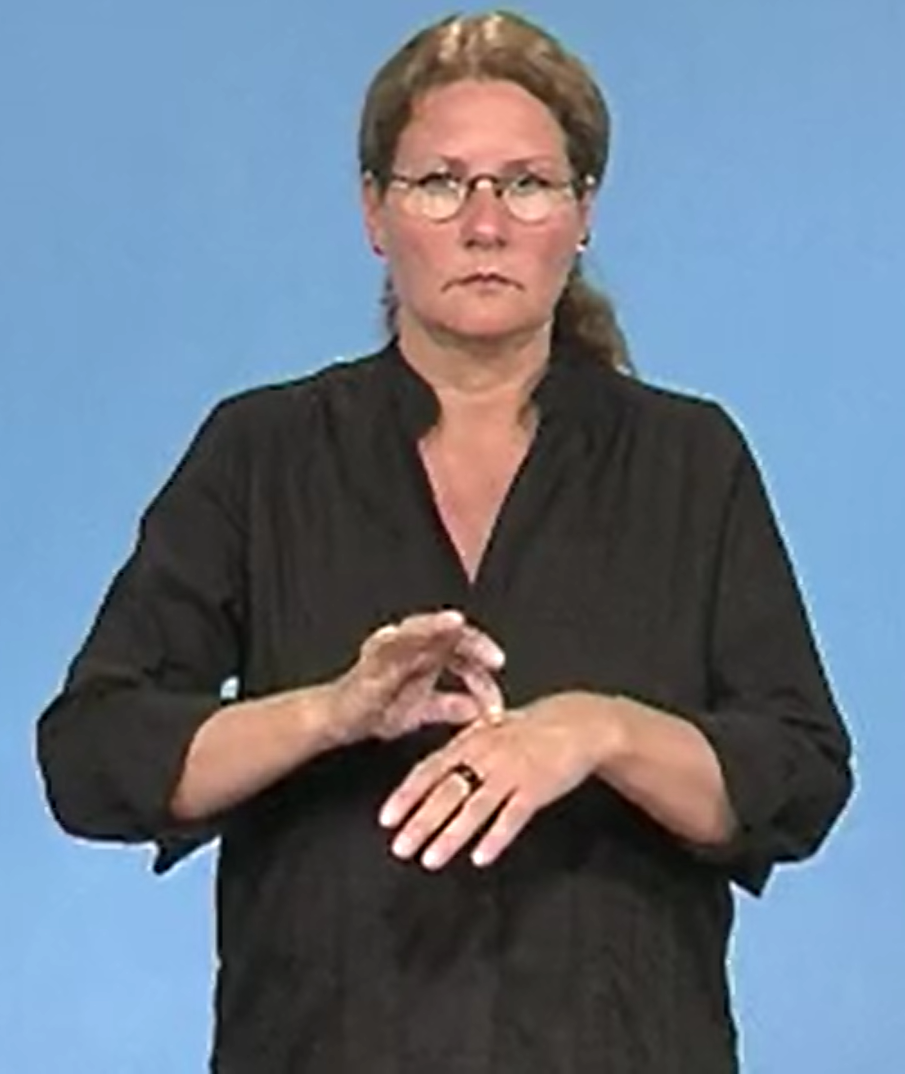} &
      
      \includegraphics[clip,trim=20 0 0 0,align=c,height=1.8cm,width=1.3cm,frame]{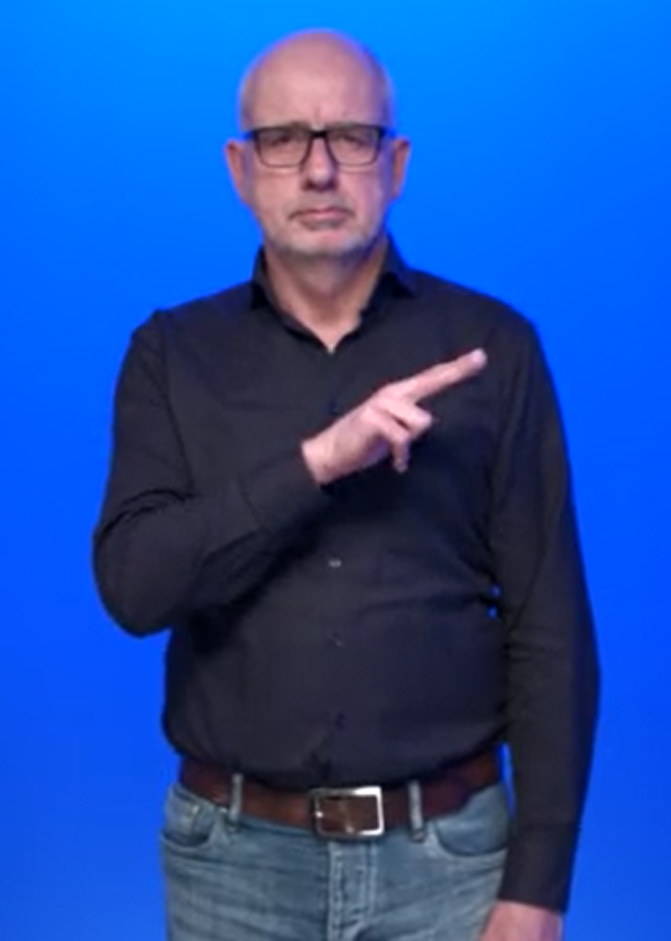} &
      
      \includegraphics[clip,align=c,height=1.8cm,width=1.33cm,frame]{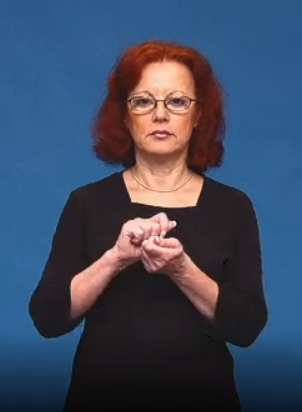} &
      
      \includegraphics[clip,align=c,height=1.8cm,width=1.33cm,frame]{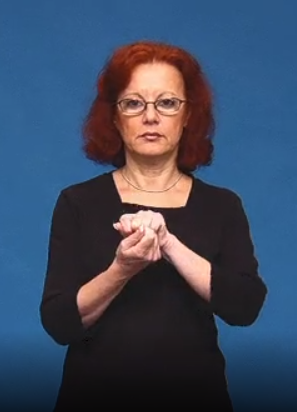} \\[0.8cm]
  
      \rotatebox[origin=c]{90}{SigML avatar} &
      \includegraphics[trim=0 89 0 0,clip,height=2.3cm,align=c,frame]{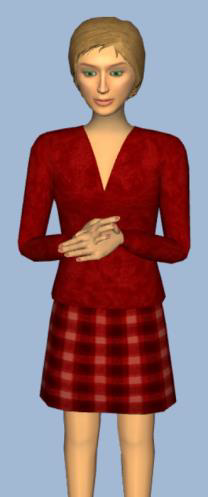} & 
      \includegraphics[trim=0 89 0 0,clip,height=2.3cm,align=c,frame]{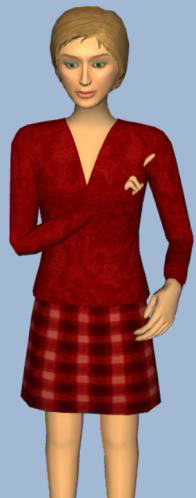} &
      \includegraphics[trim=0 135 0 0,clip,height=2.3cm,align=c,frame]{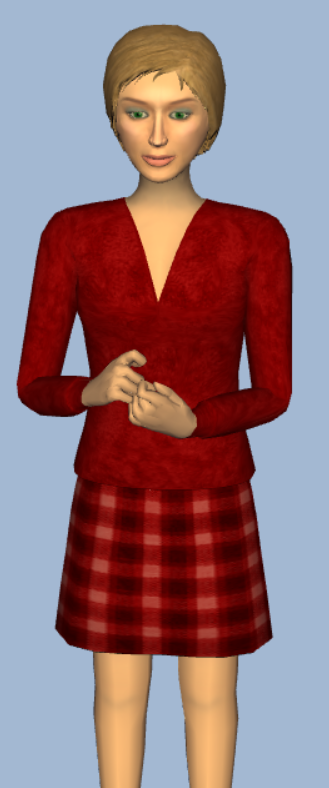} &
      \includegraphics[trim=0 135 0 0,clip,height=2.3cm,align=c,frame]{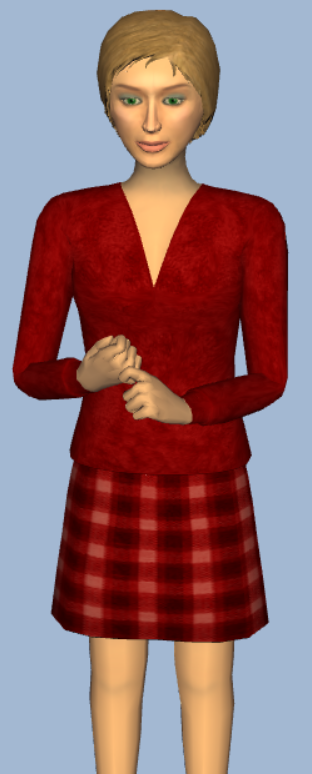} \\
    \end{tabular}
    
  \caption{\textbf{JASign (SigML) failure cases.} hand-inside-hand \\ (H-in-H), hand-inside-clothes (H-in-C) artifacts, wrong signing.
  }
  \label{fig:failure_cases}
  \vspace{-0.2cm}
\end{figure}

Since the early 2000s, there have been several research projects exploring avatars animated from HamNoSys, such as VisiCast \cite{bangham2000virtual}, eSign \cite{zwitserlood2004synthetic}, dicta-sign \cite{efthimiou2010dicta}, and JASigning \cite{ebling2016building}. While these avatars produce sign sequences, they are not popular among the deaf community due to under-articulated and unnatural movements, making the avatars difficult to understand  \cite{stoll2020text2sign}.
Furthermore, the robotic movements of these avatars can make viewers uncomfortable due to the uncanny valley phenomenon \cite{mori2012uncanny}. In addition, as illustrated in \Cref{fig:failure_cases}, these avatars do not perform all hand motions correctly. 
A later work \cite{gibet2016interactive}, uses motion capture data to create more stable and realistic avatars. However, this method is limited to a small set of phrases due to the high data collection and annotation cost.

\subsection{Gloss Approaches}
\label{subsec:glosses}
To cope with these challenges, a recent work \cite{stoll2020text2sign} suggests combining generative models with a motion graph (MG) \cite{kovar2008motion} and Neural Machine Translation. They translate spoken language sentences into gloss sequences that condition an MG to find a pose sequence representing the input from a dictionary of poses. The sequence is then converted to a video using a GAN.
A similar work \cite{saunders2020progressive} suggests progressive transformers for generating signed pose sequences from spoken language through glosses. Like Stoll~\etal \cite{stoll2020text2sign}, they use a closed set of dictionary signs as the signs in their output sequence, which makes these solutions language and data-specific.
A later work \cite{saunders2021mixed} uses learned ``cheremes\footnote{A Sign language equivalent of phonemes.}'' to generate signs. Similarly to glosses and phonemes, cheremes are language specific.
In contrast, since our method uses a universal notation, it works for languages it was trained on and for unseen languages, as long as the individual glyphs exist in our dataset.

\subsection{HamNoSys Generation}
\label{subsec:hamnosys_generation}
Recently, different HamNoSys generation tasks have been researched. For example, Skobov~\etal suggested a method for automatically annotating videos into HamNoSys \cite{skobov2020video}. Further research on this task could enhance the capabilities of our model, by creating more labeled data. Translating spoken language text into HamNoSys has also been researched  \cite{walsh2022changing}, and if improved, can allow a complete translation pipeline from spoken language text into Sign languages using our model.

\subsection{Diffusion Models}
\label{subsec:diffusion}

Diffusion models \cite{sohl2015deep, ho2020denoising} recently showed impressive results on image and video generation tasks \cite{dhariwal2021diffusion, ramesh2022hierarchical, saharia2022photorealistic}. Generation is done using a learned gradual process, with equal input and output sizes. The model gradually changes the input to get the desired output.
In this work, we take inspiration from diffusion models in the sense that our model learns to gradually convert the input (a sequence of a duplicated reference frame) into the desired pose sequence.

\subsection{Text to Motion}
\label{subsec:text2motion}

In recent years, works on motion generation from English text \cite{ahuja2019language2pose, ghosh2021synthesis, petrovich2022temos, tevet2022human, guo2022generating} showed impressive results. While these works may seem related to our task, they use 3D motion capture data, which is not available for our task. As detailed in \Cref{sec:data}, our data is collected using a pose estimation model over sign videos; thus, it is both 2D and imperfect, with many missing and incorrect keypoints.
Moreover, since the text in these works is written in English, recent works \cite{ghosh2021synthesis, petrovich2022temos, tevet2022human} take advantage of large pre-trained language models such as BERT \cite{devlin2018bert}, CLIP \cite{radford2021learning}, etc. As HamNoSys is not a common language, with limited available resources, we cannot use pre-trained models as they do.
\section{Data}
\label{sec:data}
\begin{figure*}[t]

    \centering

    \setlength{\tabcolsep}{0.1pt}

    \renewcommand{\arraystretch}{0.5}

    \resizebox{.9\textwidth}{!}{
    \begin{tabular}{c@{\hskip 0.6em}c@{\hskip 0.1em}c@{\hskip 0.1em}c@{\hskip 0.1em}c@{\hskip 0.3em}c@{\hskip 0.1em}c@{\hskip 0.1em}c@{\hskip 0.23em}c}

        \rotatebox{90}{\phantom{Aaa.}{video}} &

        \includegraphics[trim=25 17 20 19, clip,width=18.6mm]{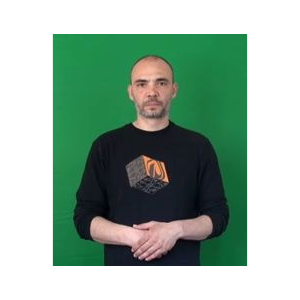} &

        \includegraphics[trim=25 17 20 19, clip,width=18.6mm]{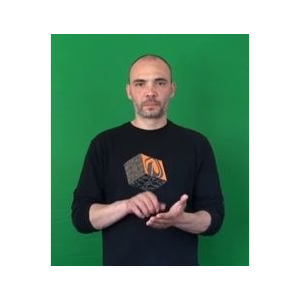} &

        \includegraphics[trim=24 17 20 18, clip,width=18.6mm]{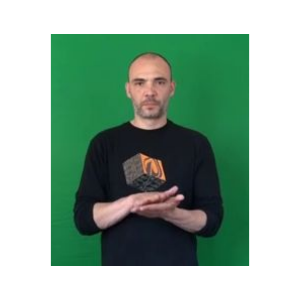} &

        \includegraphics[trim=25 17 20 19, clip,width=18.6mm]{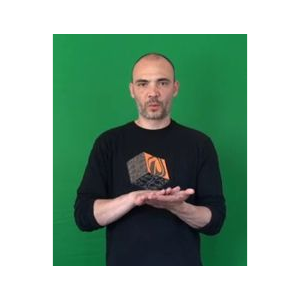} &
        
        \includegraphics[trim=0 10.5 0 0,clip,width=18mm]{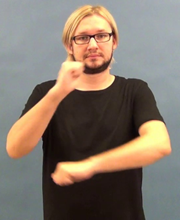} &

        \includegraphics[trim=0 12.5 0 0,clip,width=18mm]{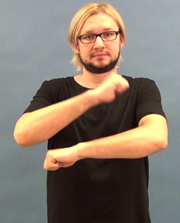} &

        \includegraphics[trim=0 12.5 0 0,clip,width=18.1mm]{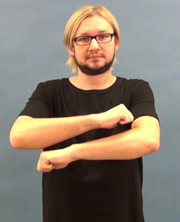} &
        
        \includegraphics[trim=0 2.5 0 3,clip,width=17.5mm]{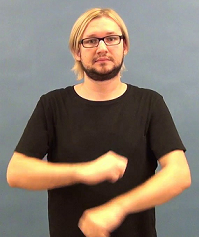} 
        \\

        \\

        \rotatebox{90}{\phantom{Aa}{original pose}} &

        \includegraphics[clip,width=16mm]{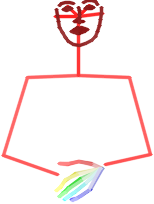} &

        \includegraphics[clip,width=16mm]{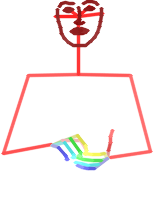} &

        \includegraphics[clip,width=16mm]{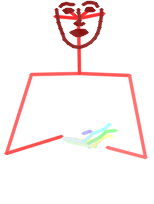} &

        \includegraphics[clip,width=16mm]{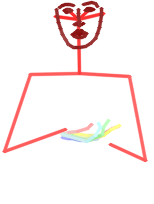} &
        
        \includegraphics[trim=0 20 0 0,clip,width=19mm]{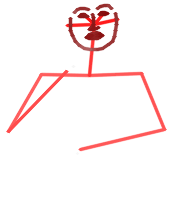} &

        \includegraphics[trim=0 20 0 0,clip,width=19mm]{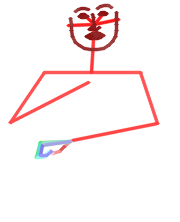} &

        \includegraphics[trim=0 20 0 0,clip,width=19mm]{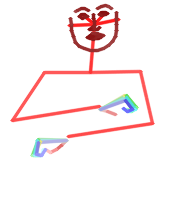} &
        
        \includegraphics[clip,width=17mm]{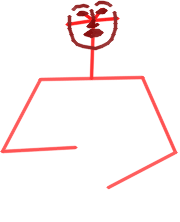} 

        \\

       \rotatebox{90}{\phantom{A}{generated pose}} &

        \includegraphics[clip,width=16mm]{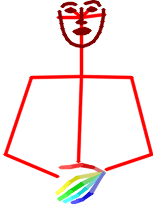} &

        \includegraphics[clip,width=16mm]{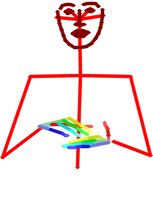} &

        \includegraphics[clip,width=16mm]{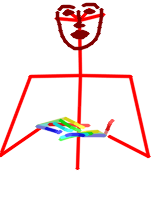} &

        \includegraphics[clip,width=16mm]{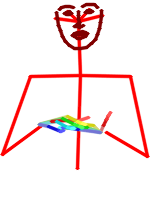} &
        
        \includegraphics[clip,width=14.8mm]{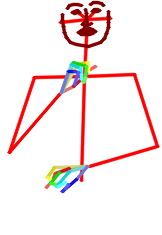} &

        \includegraphics[clip,width=15mm]{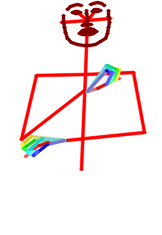} &

        \includegraphics[clip,width=15mm]{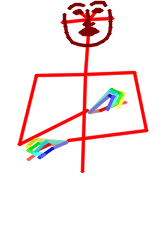} & 

        \includegraphics[clip,width=15.4mm]{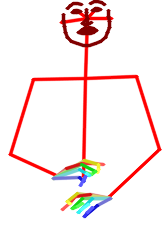}

    \end{tabular}
    }
    \caption{\textbf{Results examples:} \textbf{Top row:} original video frames, \textbf{middle row:} ground truth pose detected by OpenPose, \textbf{bottom row:} generated pose. Despite missing keypoints in the ground truth pose, our model generates a correct pose.}
    \label{fig:results_example}
    \vspace{-0.3cm}
\end{figure*}

Our dataset consists of $5,754$ videos of Sign languages signs with their HamNoSys transcriptions. Each video is of a front-facing person signing a single sign.
We collect the data from the DGS Corpus \cite{prillwitz2008dgs}, Dicta-Sign \cite{matthes2012dicta}, and the corpus-based dictionary of Polish Sign Language \cite{linde2014corpus}. Together, the data contains four Sign languages, signed by 14 signers:
Polish SL (PJM): 2,560 signs, 2 signers; German SL (DGS): 1,926 signs, 8 signers; Greek SL (GSL): 887 signs, 2 signers; and French SL (LSF): 381 signs, 2 signers.

\subsection{Data Pre-Processing}
\label{subsec:preprocess}
To use the collected data as ground truth (GT) for sign pose sequence generation, we extract estimated pose keypoints from each video using the OpenPose \cite{cao2017realtime} pose estimation model.
Each keypoint $k_i\in{K}$ consists of a 2D location $(x, y)$ and the confidence of the model, $c_i$. Missing keypoints (keypoints with $c_i=0$) or ones with a confidence of less than $20\%$ are filtered out.
We further process the extracted keypoints as follows:

\begin{enumerate}[leftmargin=*]
    \item \textbf{Trim meaningless frames.} 
    Some videos (\eg videos fading in and out) contain leading or trailing frames that do not contain enough relevant information. Moreover, in resting position, hands are not always visible in the video frame. Hence, leading / trailing frames in which the face or both hands are not identified are removed.
    \item \textbf{Mask legs and unidentified keypoints.}
    In addition to setting a confidence threshold, since the legs are not significant, we remove them by setting the confidence of every keypoint from the waist down to 0, allowing the model to only learn from existing and relevant keypoints.
    \item \textbf{Flip left-handed sign videos.} One-hand signs are usually signed with the dominant hand. A left-handed person would mirror a sign signed by a right-handed person. Given that the signing hand is not specified in HamNoSys and that some videos in our dataset include left-handed signers, for consistency, we flip these videos to produce right-handed signs.
    \item \textbf{Pose Normalization.} We normalize the pose keypoints by the pose shoulders, using the \emph{pose\_format} library \cite{moryossef2021pose-format}, so all poses have the same scale. It defines the center of a pose to be the neck, calculated as the average middle point between the shoulders across all frames. Then, it translates all keypoints, moving the center to $(0, 0)$, and scales the pose so the average distance between the shoulders is 1.
\end{enumerate}

We note that the data is still imperfect, with many missing and incorrect keypoints, as seen in \Cref{fig:results_example}.

\section{Method}
\label{sec:method}
\begin{figure*}[htpb]
  \centering
   \includegraphics[width=\textwidth]{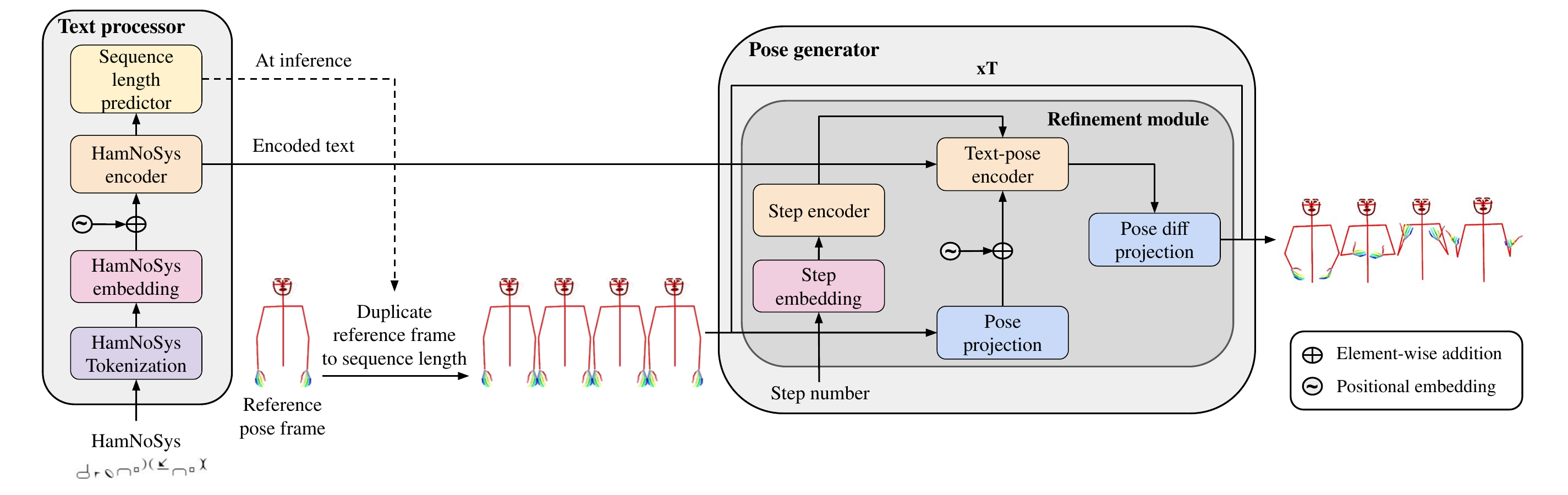}
   \caption{\textbf{Model architecture.} First, the text processor encodes the HamNoSys and predicts the sequence length. Next, the reference pose is duplicated to the sequence length and passed to the pose generator, which iteratively uses the current pose sequence and HamNoSys encoding for T steps and generates the desired pose. After T steps, the pose generator outputs the final pose sequence.}
   \label{fig:arch}
\end{figure*}

Given a sign written in HamNoSys, our model generates a sequence of frames signing the desired sign.
We start from a single given ``reference'' pose frame, acting as the start of the sequence, and duplicate it to the length of the signed video.
The sign is generated gradually over $T$ steps, where in each time step $t\in{\{T\dots 0\}}$ the model predicts the required change from step $t$ to step $t-1$ as described in \Cref{subsec:refinement}. 
Our method takes inspiration from diffusion models in that it learns to gradually generate a sign pose from a reference pose and a HamNoSys notation guidance.
Unlike diffusion models, we start from a reference pose and not from random noise to allow for the continuation of signs with the same identity. Furthermore, at each prediction step, we predict the change required to move to the previous step rather than predicting the change required to move to the final sequence. This way, the model can replace missing and incorrect keypoints with correct ones in a gradual process. We combine these ideas with transformer encoders \cite{vaswani2017attention} for the pose and HamNoSys, leading to meaningful representations, considering their spatial and temporal meanings.

\subsection{Model Architecture}
\label{subsec:model_arch}
As shown in \Cref{fig:arch}, the model is composed of two parts: the text processor (\Cref{subsec:text_proc}), responsible for the HamNoSys text encoding and predicting the length of the generated pose sequence; and the pose generator (\Cref{subsec:pose_gen}), responsible for the pose sequence generation.
The inputs to the model are a HamNoSys sequence and a single pose frame. The reference frame is the starting point of the generated pose. It can either be a resting pose or the last frame of a previously generated sign for continuity purposes.
The pipeline of the model is as follows: the input text is tokenized, embedded, and encoded by the text processor; then, it is passed to the sequence length predictor that predicts the sequence length. Next, the reference pose is duplicated to the desired pose length (during training, it is the GT length, while at inference, it is the length predicted by the sequence length predictor) and is passed to the pose generator with the encoded text. 
Finally, using the encoded text and extended reference pose, the pose generator (\Cref{subsec:pose_gen}) gradually refines the pose sequence over $T$ steps to get the desired pose. This process is summarized in \Cref{alg:text2pose}.

\subsection{Text Processor}
\label{subsec:text_proc}
This module is responsible for the HamNoSys text processing.
The HamNoSys text is first tokenized into a tokens vector, so each glyph gets a unique identifier (token). Next, the tokens' vectors are passed through a learned embedding layer, producing vector representations for each token.
In addition, we use a learned embedding layer as a positional embedding to represent the positions of the sequence tokens, so the model gets information about the order of the tokens. The vector representations of the tokens' locations are of the same dimension $D$ of the tokens' vector representations, to allow the summation of them.
After the tokens and their locations are embedded and combined, they are passed to the HamNoSys transformer encoder \cite{vaswani2017attention}.
Finally, the encoded text is passed to the pose generator and to the sequence length predictor--- a linear layer that predicts the length of the pose sequence.

\subsection{Pose Generator}
\label{subsec:pose_gen}
\begin{figure}[t]
    \centering
    \setlength{\tabcolsep}{0.5pt}
    \renewcommand{\arraystretch}{0.5}

\begin{adjustbox}{max width=\linewidth}
    \begin{tabular}{cccccc}

        Image &
        Pose &
        Add &
        Replace &
        Blend \\
        
        \includegraphics[trim=5 15 0 0,clip,width=16mm]{assets/pjm_1685/pjm_1685_vid_2.png} &
        
        \includegraphics[clip,width=15mm]{assets/pjm_1685/pjm_pose0_15.png} &

        \includegraphics[clip,width=15mm]{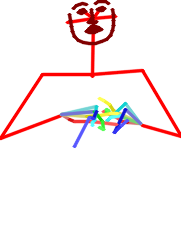} &
        
        \includegraphics[clip,width=16mm]{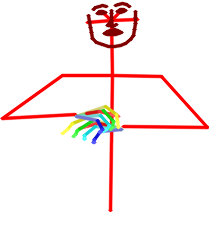} &
        
        \includegraphics[clip,width=13mm]{assets/pjm_1685/pjm_pose1_15.png} 

    \end{tabular}
\end{adjustbox}
    \caption{\textbf{Blend importance example.} Left to right: original image, original pose, pose generated by addition, by replacement, and by blend.}
    \label{fig:no_blend_example}
    \vspace{-0.2cm}
\end{figure}

The pose generator is responsible for the sign pose sequence generation, which is the output of the entire model. It does so gradually over $T$ steps, where at time step $T$ ($s_T$), the sequence is the given reference frame extended to the sequence length.
During training, this is the actual length of the sign video after frame trimming (\Cref{subsec:preprocess}), while at inference, this is the length predicted by the sequence length predictor.
To generate the desired sign gradually, we define a schedule function ($\delta \in [0,1]$) as $\delta_t = log_T{(T-t)}$, a step size $\alpha_t = \delta_{t} - \delta_{t+1}$\footnote{for $t=T-1$ we use a constant 0.1 to avoid illegal calculations}
and the predicted pose sequence at time step $t$, $\hat{s_t}$, for $t\in\{T-1,\dots,0\}$, as:
\begin{equation}
\label{eq_st}
\hat{s_t} = \alpha_t p_t + (1 - \alpha_t) \hat{s_{t+1}}
\end{equation}

where $p_t$ is the pose value predicted by the refinement module (\Cref{subsec:refinement}) at time step $t$.
This way, since the previous step result is input to the current step, and the step size decreases over time, the model needs to predict smaller changes in each step. 
As a result, the coarser details are generated first, and the finer details are generated as the generation process proceeds.
Moreover, since the result of each step is a blending between the previous step result and the current prediction and not only an addition of the prediction to the previous step, we give less weight to the initial pose sequence at each step. This way, the model can fill missing and incorrect keypoints by gradually replacing them with correct data.
Additionally, since it is a gradual process, where the model predicts small changes at each step instead of predicting and replacing the whole pose sequence, its results are more smooth and accurate. \Cref{fig:no_blend_example} shows the importance of blending.
To make the model more robust, we add $\epsilon z$ noise to $\hat{s_t}$ ($z\sim N(0,I)$) at each time step during training.
Finally, $s_0$ is returned as the sign pose prediction.

\subsubsection{Refinement Module}
\label{subsec:refinement}

At each step $t\in\{T-1,\dots,0\}$, the pose generator calls the refinement module with the previously generated pose $\hat{s_{t+1}}$, the pose positional embedding, the step number, and the encoded text.
The step number is passed through an embedding layer that produces a vector representation of dimension $D$, which is encoded using two linear layers with activations between them.
Similarly, each pose frame of $\hat{s_{t+1}}$ is projected to a vector representation of dimension $D$, using two linear layers with activation. 
The projected result is summed with the positional embedding of the pose to form a pose embedding. The pose embeddings are then concatenated with the encoded text and step, and together, as shown in \Cref{fig:arch}, they are passed to the text-pose transformer encoder \cite{vaswani2017attention}. 
Finally, the result is passed to the ``\emph{pose diff projection}'', formed of two linear layers with activation, which generates the predicted pose for the current step, $p_t$, that is the output of this module. \newline

\begin{algorithm}
	\caption{text2pose} 
	\hspace*{\algorithmicindent} 
	\textbf{Input:} \textbf{text:} HamNoSys tokens \\
	\hspace*{16mm} 
	\textbf{ref\_pose:} a single reference pose frame \\ 
 \hspace*{\algorithmicindent} \textbf{Output:} $\mathbf{s_0}$\textbf{:} pose sequence prediction 
	\begin{algorithmic}[1]
		\State $et, seq\_len = process\_text(text)$
		\State $s_T = [ref\_pose]*seq\_len $
		\State $pe = positional\_embedding([0,\dots,seq\_len]) $
		\For {$step=T-1,\dots,0$}
			\State $ se = encode\_step(step)$
		    \State $p_t = refine(s_{t+1}, et, pe, se) $
			\State $s_t = \alpha_t p_t + (1-\alpha_t)s_{t+1} + \epsilon z$
		\EndFor
	\end{algorithmic} 
	\label{alg:text2pose}
\end{algorithm}

\subsection{Loss Function}
\label{subsec:loss_func}

At every refinement step we want to compare the predicted sequence to an interpolation between the real sequence $s_0$ and the starting sequence $s_T$. However, as our model uses the prediction of the previous step, at time step $t$ we interpolate between $s_0$ and the previous step $s_{t+1}$. For that purpose, we define:
\begin{equation}
s_t=\delta_{t} s_0 + (1 - \delta_{t}) s_{t+1}
\label{eq:gt_st}
\end{equation}

We mark the $i$th joint in the $j$th frame in $s_t$ by $s_t^i[j]$.
The refinement loss function $L_{p}$ at time step $t$ is a weighted MSE with weight $c_i[j]$ for each joint $i\in{K}$ in frame $j\in\{0,\ldots,N\}$:

\begin{equation}
L_{p}(s_t, \hat{s_t}) =
\frac{1}{N}\sum_{j=0}^N{\frac{1}{|K|}\sum_{i=0}^{|K|}{c_i[j](s_t^i[j]-\hat{s_t^i[j]})^2}}
\end{equation}

To avoid affecting the learning rate when experimenting with different step numbers, we scale the loss by $ln(T)^2$ (See full derivation in \Cref{supp:loss_deriv}).

To train the sequence length predictor, we calculate MSE loss between the predicted sequence length $\hat{N}$ and the real one $N$, and add it to the final loss with a small weight of $\gamma$. The complete loss term is then:

\begin{equation}
L = ln(T)^2 L_{p}(s_t, \hat{s_t}) + \gamma \cdot L_{len}(N, \hat{N})
\end{equation}

\section{Implementation Details}
\label{sec:implementation_details}

We provide a detailed model architecture in \Cref{supp:arch}.
We use learned embedding layers with dimension $D=128$ to embed the HamNoSys text, the step number, and the text and pose positions.
For the pose and text encoding layers, we use a Transformer encoder \cite{vaswani2017attention} with two heads, with depths 2 and 4 for the text and pose respectively.
We project the pose frames using two linear layers with swish activation \cite{ramachandran2017Swish} between them to a vector with the same dimension $D=128$ for each frame. The step encoder and the pose diff projection are also formed of two linear layers with swish activations between them.
After experimenting with three step number options, we set $T = 10$ as the number of steps for the pose generation process in all our experiments. 
We discuss the experiments and their results in \Cref{sec:ablations}.
We train using the Adam Optimizer for 2000 epochs, setting the learning rate to 1e-3 empirically, and teacher forcing
with probability of 0.5. For the noise addition we use $\epsilon=1e-4$, and for the sequence length loss weight we use $\gamma=2e-5$. We use this value because a lower value prevents the sequence length predictor from learning, while a higher value prevents the refinement module from learning. \newline

\subsection{Teacher Forcing}
\label{subsec:teacher_forcing}
During training, we use teacher forcing \cite{williams1989learning} in the pose generation process.
with a probability of $p=0.5$. 
Meaning, with a probability of $p$, we feed the refinement module (\Cref{subsec:refinement}) with $s_t$ as defined in \Cref{eq:gt_st}, and with probability of $1-p$ we feed it with the predicted $\hat{s_t}$ as defined in \Cref{eq_st}, to help the model learn faster.
Hence, during training, $s_t$ is defined by:

\begin{equation}
\begin{aligned}
&f\sim Ber(p); z\sim N(0,I) \\
&s_t = \begin{cases}
\alpha_t p_t + (1 - \alpha_t) s_{t+1} + \epsilon z & f = 0 \\
\delta_{t} s_0 + (1 - \delta_{t}) s_{t+1} + \epsilon z             & \text{otherwise}
\end{cases}
\end{aligned}
\label{eq_st_train_tf}
\end{equation}

\section{Evaluation}
Currently, there is no suitable evaluation method for SLP in the literature. 
APE (Average Position Error) is a distance measurement used to compare poses in recent text-to-motion works \cite{ahuja2019language2pose, ghosh2021synthesis, petrovich2022temos}. 
It is the average L2 distance between the predicted and the GT pose keypoints across all frames and data samples. Since it compares absolute positions, it is sensitive to different body shapes and slight changes in timing or position of the performed movement.
Huang~\etal~\cite{huang2021towards} suggested \emph{DTW-MJE} (Dynamic Time Warping - Mean Joint Error), which measures the mean distance between pose keypoints after aligning them temporally using \emph{DTW} \cite{kruskal1983overview}. However, it is unclear from the original \emph{DTW-MJE} definition how to handle missing keypoints, hence we suggest a new distance function that considers missing keypoints and apply \emph{DTW-MJE} with our distance function over normalized keypoints. We mark this method by \emph{nDTW-MJE}.
We validate the correctness of \emph{nDTW-MJE} by using AUTSL \cite{sincan2020autsl}, a large-scale Turkish Sign Language dataset, showing that it measures pose sequences distance more accurately than existing measurements.
Then, we evaluate the results of our model using \emph{nDTW-MJE} in two ways:
Distance ranks (\Cref{subsec:ranks}) and Leave-one-out (\Cref{subsec:leave_one_out}).
We test the sequence length predictor using absolute difference between the real and predicted sequence length. The mean difference is 3.61, which usually means more resting pose frames in one of the poses. 
Finally, we provide qualitative results in \Cref{fig:results_example} and on the project page.

\subsection{Distance measurement (\emph{nDTW-MJE})} 
\label{subsec:eval_method}
\begin{table}
  \adjustbox{max width=1.0\linewidth}{
  \centering
  \begin{tabular}{@{}c@{\hskip 1em}c@{\hskip 1em}c@{\hskip 1em}c@{\hskip 1em}c}
    \toprule
    Method & prec@1 $\uparrow$ & prec@5 $\uparrow$ & prec@10 $\uparrow$ & mAP $\uparrow$ \\
    \midrule
    MSE & 0.5 & 0.36 & 0.34 & 0.27 \\
    nMSE & 0.82 & 0.68 & 0.62 & 0.4 \\
    APE & 0.82 & 0.55 & 0.45 & 0.29 \\
    nAPE & 0.93 & 0.78 & 0.7 & 0.44 \\
    DTW-MJE & 1 & 0.78 & 0.66 & 0.33 \\
    nDTW-MJE & \textbf{1} & \textbf{0.9} & \textbf{0.84} & \textbf{0.58} \\
    \bottomrule
  \end{tabular}
  }
  \caption{Distance measurements results over AUTSL}
  \label{tab:autsl}
  \vspace{-0.2cm}
\end{table}
We suggest a new method for measuring the distance between two pose sequences using \emph{DTW-MJE} \cite{huang2021towards} over normalized pose keypoints, that considers missing keypoints.
To measure the distance between two pose sequences, we use a variant of \emph{DTW} \cite{kruskal1983overview}. \emph{DTW} is an algorithm that can measure the distance between two temporal sequences, because it is able to ignore global and local shifts in the time dimension.
We use a linear-time approximation of \emph{DTW} suggested by salvador~\etal~\cite{salvador2007toward} while considering missing keypoints using the following distance function for each keypoints pair:

\begin{equation} \label{eq:dtw_dist}
\begin{split}
& dist(ref, other) = \\
& \begin{cases}
    0 & ref = NaN \\
    \frac{\norm{ref}_2}{2} & ref \neq NaN \quad \& \quad other = NaN \\
    \norm{ref-other}_2 & \text{otherwise}
\end{cases}
\raisetag{4\baselineskip}
\end{split}
\end{equation}

This way, we only compare keypoints existing in both poses and punish keypoints that exist in the $reference$ pose but not in the $other$ pose.
\newline

To measure the distance from $ref$ to $other$, we normalize them as described in \Cref{sec:data} 
and calculate the \emph{DTW-MJE} distance using \Cref{eq:dtw_dist} between the normalized pose sequences.\\

\subsection{nDTW-MJE Validation}
\label{subsubsec:validate_eval_method}
To validate \emph{nDTW-MJE} with our suggested distance measurement, we test it against AUTSL \cite{sincan2020autsl} --- a large-scale Turkish Sign Language dataset comprising 226 signs performed by 43 different signers in different positions and postures and 38,336 isolated sign video samples in total.
We collect all samples for every unique sign $s$, and randomly sample $4 \times |s|$ other signs from the dataset to keep a $1:4$ ratio of $same:other$ samples. 
Then, we measure the distance from the reference sign to each one of the samples using \emph{nDTW-MJE} as explained in \Cref{subsec:eval_method}, and calculate \emph{mAP} (mean Average Precision) and \emph{mean precision@k} for $k=1, 5, 10$, across all signs, measuring how many of the $k$ most similar poses were of a video signing the query sign. 
In \Cref{tab:autsl}, we compare the results of using \emph{nDTW-MJE} vs. using \emph{MSE} and \emph{APE} over normalized and unnormalized pose keypoints and of using \emph{DTW-MJE} over unnormalized pose keypoints (with our distance function).
The results show that \emph{DTW-MJE} with our distance function better suits pose sequences distance measurement than the more commonly used \emph{APE}. Moreover, using normalized keypoints improves the performance of all measurements, while \emph{nDTW-MJE} measures the distances more accurately than all other options.

\subsection{Distance Ranks}
\label{subsec:ranks}
\begin{table}
  \adjustbox{max width=1.0\linewidth}{
  \centering
  \begin{tabular}{@{}c@{\hskip 1.5em}c@{\hskip 1.5em}c@{\hskip 1.5em}c@{\hskip 1.5em}c}
    \toprule
    Reference & Model & Rank 1 $\uparrow$ & Rank 5 $\uparrow$ & Rank 10 $\uparrow$ \\
    \midrule
    \multirow{3}{1.7cm}{\centering Prediction} & PT & 0.04 & 0.19 & 0.34 \\
    & Resting & 0.05 & 0.17 & 0.31 \\
    & \textbf{Ours} & \textbf{0.08} & \textbf{0.2} & \textbf{0.35} \\
    \midrule
    \multirow{3}{1.7cm}{\centering Ground Truth} & PT & 0.0 & 0.001 & 0.02 \\
    & Resting & 0.008 & 0.08 & 0.16 \\
    & \textbf{Ours} & \textbf{0.21} & \textbf{0.44} & \textbf{0.56} \\
    \bottomrule
  \end{tabular}
}
  \caption{\textbf{Distance ranks.} Top: distance to prediction. Bottom: distance to GT pose. We compare our results to the Progressive Transformers (PT), and to a sequence of constant resting positions. }
  \label{tab:dist2both}
\end{table}
To evaluate our animation method, we calculate the distance between each prediction and GT pair using \emph{nDTW-MJE}.
Then, for each pair, we create a gallery of 20 random GT samples from the dataset and 20 predictions for other samples from the dataset. For each gallery sample, we calculate its distance from the prediction and the GT as references.
Finally, we calculate rank $1$, $5$, and $10$, where rank $k$ is the percentage of test samples for which the tested pose (GT pose for the distance to prediction case, prediction for the distance to GT case), was in the $k$ most similar poses to the reference pose and report them in \Cref{tab:dist2both}.
For comparison, we use the most similar previous work to ours, Progressive Transformers (PT) \cite{saunders2020progressive}, which aims to translate glosses into pose sequences. We adjust their model to take HamNoSys sequences as input instead of glosses and train it over our data.
We also present the results of using $s_T$ of each GT (\ie a sequence of the first pose frame of the GT in the length predicted by the sequence length predictor) as the prediction for comparison. 
As shown, our model outperforms both PT and the ``resting" option in both settings.

\subsection{Leave One Out}
\label{subsec:leave_one_out}
\begin{table}
\adjustbox{max width=1.0\linewidth}{
  \centering
  \begin{tabular}{@{}c@{\hskip 1em}c@{\hskip 2em}c@{\hskip 2em}c@{\hskip 2em}c}
    \toprule
    Ref. & Lang. & Rank 1 $\uparrow$ & Rank 5 $\uparrow$ & Rank 10 $\uparrow$ \\
    \midrule
    \multirow{4}{1.7cm}{\centering Pred.} 
    & PJM & 0.03 / 0 & 0.15 / 0.005 & 0.27 / 0.02 \\
    & DGS & 0.14 / 0.02 & 0.23 / 0.08 & 0.39 / 0.14 \\
    & GSL & 0 / 0.01 & 0.1 / 0.06 & 0.25 / 0.14 \\
    & LSF & 0.22 / 0.18 & 0.6 / 0.41 & 0.8 / 0.58 \\
    \midrule
    \multirow{4}{1.7cm}{\centering Ground Truth} 
    & PJM & 0.08 / 0.06 & 0.25 / 0.3 & 0.41 / 0.5 \\
    & DGS & 0.26 / 0.21 & 0.49 / 0.49 & 0.58 / 0.64 \\
    & GSL & 0.27 / 0.53 & 0.67 / 0.84 & 0.8 / 0.92 \\
    & LSF & 0.7 / 0.75 & 0.95 / 0.91 & 0.97 / 0.95 \\
    \bottomrule
  \end{tabular}
  }
  \caption{\textbf{Distance ranks by language} (full / leave-one-out model). Top: distance to prediction, bottom: distance to ground truth. }
  \label{tab:dist_by_lang}
  \vspace{-0.2cm}
\end{table}

To check if our method truly is generic, we perform a ``leave-one-out'' experiment: we train our model with all languages but one and test it on the left-out language. Then, we report rank 1, 5, and 10 results for each model in \Cref{tab:dist_by_lang}. For comparison, we also show the rank results of the full model per language.
For the leave-one-out experiment, all `other' samples are taken from the left-out language.
As demonstrated, the results when testing on unseen languages are only slightly worse than the results of the full model, and in some cases, they are even better. The degradation in results might be due to insufficient data or rare glyphs that only appear in one language.

\subsection{Qualitative Results}
\label{subsec:qualitative_results}

We present results in \Cref{fig:results_example} and in the project page. Although some frames of the GT pose often have missing or incorrect keypoints, our model generates a complete and correct pose.
\section{Limitations and Future Work}
\label{future_work}
\begin{figure}[t]

    \centering
    \setlength{\tabcolsep}{0.5pt}
    \renewcommand{\arraystretch}{0.5}

    \begin{tabular}{@{}c@{\hskip 0.5em}c@{\hskip 0.5em}c}

        Image &
        GT pose & 
        Prediction \\
        
        \includegraphics[trim=5 0 6 0 ,width=0.2\linewidth]{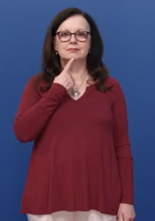} &

        \includegraphics[width=0.2\linewidth]{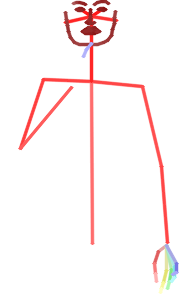} &
        
        \includegraphics[width=0.2\linewidth]{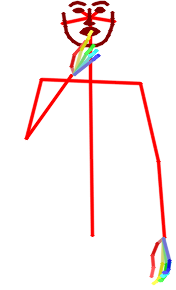} 

    \end{tabular}
    \caption{\textbf{Hand location failure case.} Pointing finger is correct in our prediction, not in the ground truth pose.}
    \label{fig:hand_loc_fail}
\vspace{-0.2cm}
\end{figure}

Despite making a big step toward SLP, much work still needs to be done.
Predicted hand shapes or movements are not always entirely correct, which we attribute to missing or incorrect keypoints in our data. The pose estimation quality is a significant bottleneck, making it almost impossible for the model to learn the correct meanings of some glyphs. \Cref{fig:results_example,fig:hand_loc_fail} show examples of missing, incorrect keypoints. Future work may improve hand pose estimation to address this issue.
Moreover, our model generates some movements correctly but not exactly in the right location due to local proximity (\eg hand over mouth instead of chin example in \Cref{fig:hand_loc_fail}. Note: the pointing finger is correct in our prediction, not in the GT pose). We present more failure examples in our project page.
Finally, more annotated data is needed. Some glyphs are rare and only appear once or a handful of times in our dataset, making it difficult for the model to learn them.

\section{Conclusion}
In this work, we propose the first method for animating HamNoSys into pose sequences. As demonstrated, our model can generate signs even when trained on partial and inaccurate data.
Additionally, we introduce a new distance function, considering missing keypoints, for measuring the distance between two pose sequences using \emph{DTW-MJE} over normalized pose keypoints.
We validate its correctness using a large-scale sign language dataset and show that it better suits pose sequences evaluation than existing methods.
We hope our method leads the way toward developing an end-to-end system for SLP, that will allow communication between the hearing and hearing-impaired communities.
\section{Acknowledgements}
\label{ack}

This project has received funding from the European Research Council (ERC) under the European Union's Horizon 2020 research and innovation programme, grant agreement No. 802774 (iEXTRACT), and from the ISF grant number 1574/21.

%%%%%%%%% REFERENCES
{\small
\bibliographystyle{ieee_fullname}
\bibliography{egbib}
}

\appendix
\section{Evaluation}
\label{supp:eval}

\subsection{Qualitative results}
\begin{figure*}

    \centering

    \setlength{\tabcolsep}{0.05pt}

    \renewcommand{\arraystretch}{0.5}

    \begin{tabular}{cccccccc}
        \raisebox{3.5\normalbaselineskip}[0pt][0pt]{\rotatebox[origin=t]{90}{original pose}} &
        \includegraphics[width=65pt]{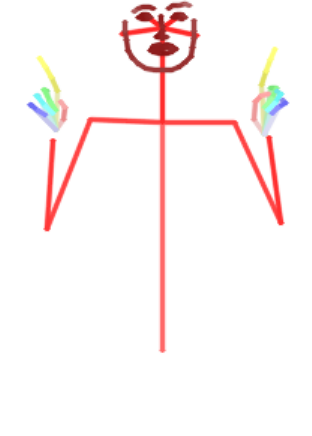} &

        \includegraphics[width=65pt]{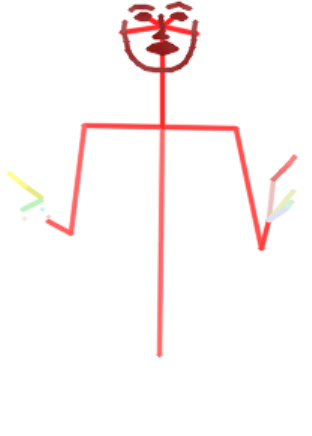} &

        \includegraphics[width=65pt]{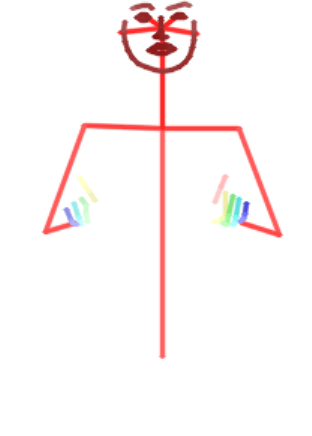} &

        \includegraphics[width=65pt]{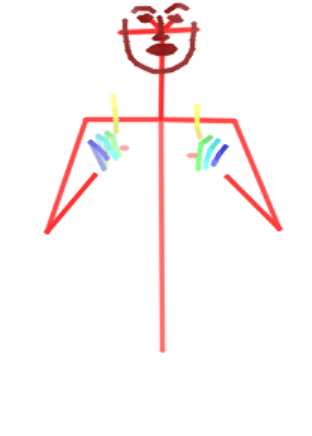} &

        \includegraphics[width=65pt]{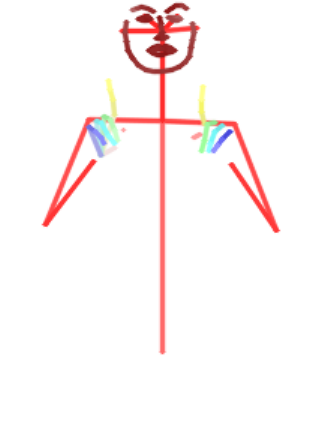} &

        \includegraphics[width=65pt]{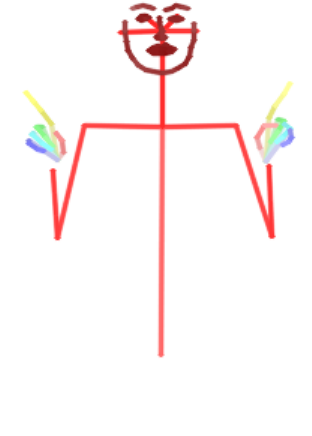} &
        
        \includegraphics[width=65pt]{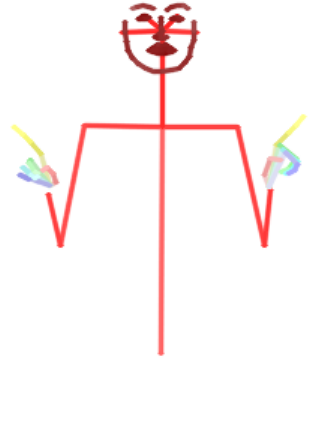}

        \\

        \raisebox{3.5\normalbaselineskip}[0pt][0pt]{\rotatebox[origin=c]{90}{generated pose}} &
        \includegraphics[width=65pt]{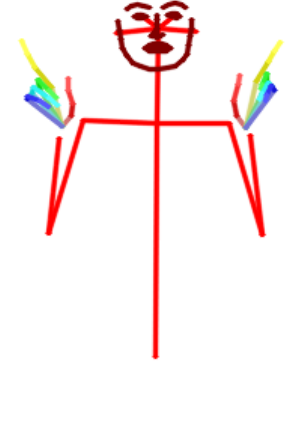} &

        \includegraphics[width=65pt]{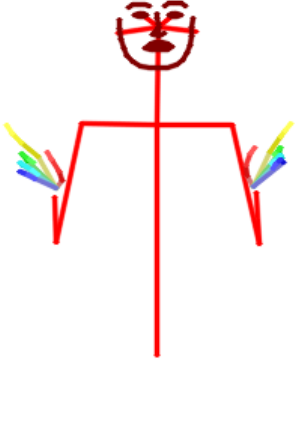} &

        \includegraphics[width=65pt]{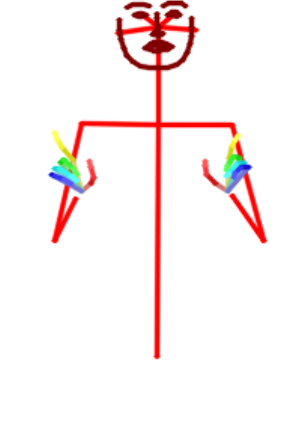} &

        \includegraphics[width=65pt]{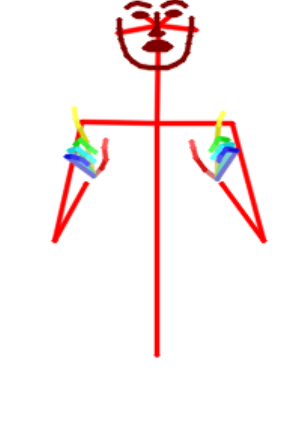} &

        \includegraphics[width=65pt]{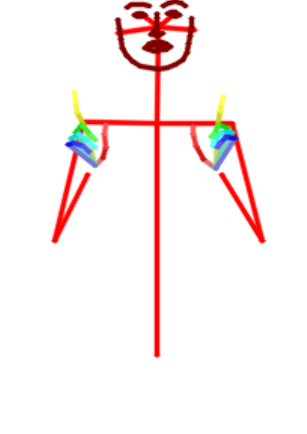} &

        \includegraphics[width=65pt]{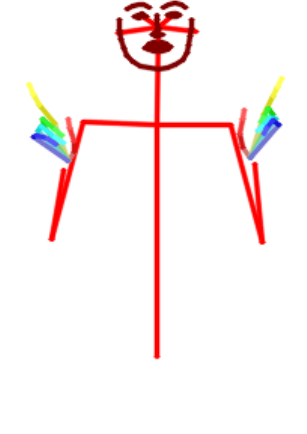} &
        
        \includegraphics[width=65pt]{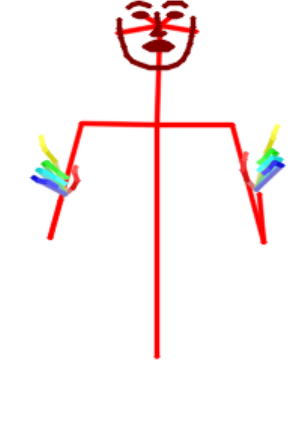}

    \end{tabular}
    \caption{\textbf{Result example.} \textbf{Top row:} ground truth pose detected by OpenPose, \textbf{bottom row:} generated pose.}
    \label{fig:sup_result_example}
\end{figure*}

We supply additional qualitative results in \Cref{fig:sup_result_example} and in the project page. We show examples of signs generated from a single HamNoSys sequence, and examples of sentences generated by concatenating a few signs generated one after the other. To generate a continuous sentence, we cut 20\% off the end of each generated sign, and give the last frame of this sequence to the model as the reference start frame for the subsequent sign. 
As demonstrated, our model is able to generate correct movements, even when the ground truth pose detected by the pose estimation model is not full or incorrect, and is able to generate multiple signs one after the other to generate sentences.

\begin{figure}
  \centering
  \includegraphics[clip, width=\linewidth, right, page=1]{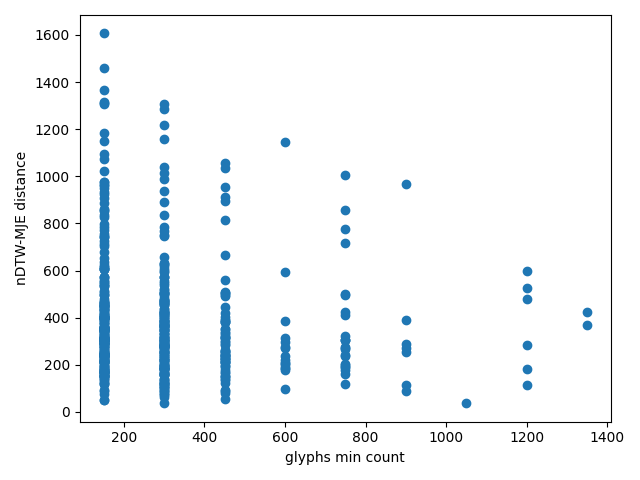}
   \caption{The number of glyph occurrences of the rarest glyph in a sequence (split into bins) vs. \emph{nDTW-MJE} distance.}
   \label{fig:num_occ2dist}
\end{figure}

\subsection{Number of glyph occurrences vs. score}
In \Cref{fig:num_occ2dist} we plot nDTW-MJE distance vs. the number of glyph occurrences of the rarest glyph per sequence, and observe that the more occurrences a glyph has, the lower the distance is. However, it is noisy, which suggests that the meaning of a glyph and the number of rare glyphs in a sequence may also affect the results.

\subsection{Sequence length}
\label{supp:seq_len}
\begin{figure}
  \centering
  \includegraphics[clip, width=\linewidth]{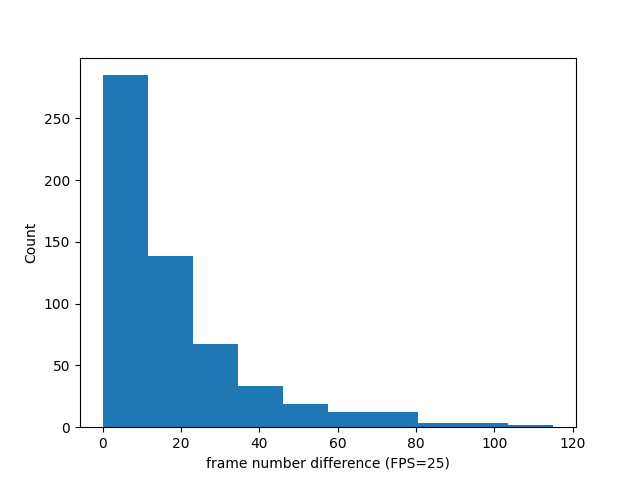}
   \caption{Absolute error between the real sequence length and the predicted one, in number of frames.}
   \label{fig:seq_len_hist}
\end{figure}

\begin{figure}
  \centering
  \includegraphics[clip, width=\linewidth]{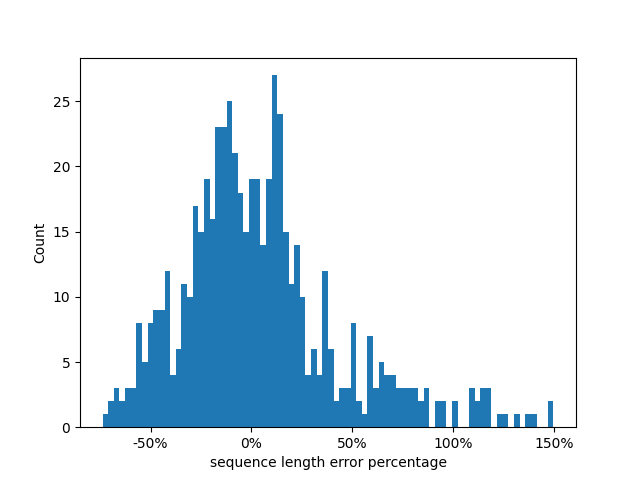}
  \caption{Signed percentage error, relative to the real sequence length, of the predicted sequence length. Negative values indicate that the predicted length is shorter than the real length; positive values indicate the opposite.}
  \label{fig:seq_len_err_hist}
\end{figure}

We test the sequence length predictor separately from the full model using absolute difference between the real sequence length and the predicted one, and show a histogram of the differences in \Cref{fig:seq_len_hist}. In addition, \Cref{fig:seq_len_err_hist} shows the percentage of error of the sequence length predictor. 
As demonstrated by the figures, the difference is low for most videos, both absolutely and relatively to the actual sequence length. Moreover, in both cases, as the sequence length difference error increases, the amount of samples with that difference decreases.

\section{Ablation studies}
\label{sec:ablations}
We conduct several ablation studies, experimenting with the amount of steps in the model, shared vs. separate positional embeddings for the text and pose, different hidden dimensions, and different feedforward dimensions. We present their results in \Cref{tab:dist2both_step,tab:ablation_pe,tab:ablation_hidden,tab:ablation_ff}.
For each of them, we train our model with the changed feature and perform the same evaluations explained in the \Cref{subsec:ranks}.

\textbf{Change in step amounts:} although the results of the 20-steps model are slightly better than the results of the 10-steps model, there is a trade-off between the number of steps and the training time of the model, as can be seen in \Cref{tab:time_per_num_steps}. Moreover, when looking at the results visually, while some results of the 20-steps model look better, others look worse. Therefore, since the improvement in the quality of the results is not drastic, we prefer to use less steps. Finally, the 5-steps model generates results that are worse both quantitatively when looking at the prediction reference, and quantitatively, hence we chose to use 10 steps in our model. We show qualitative results for each number of steps in the video on our project page.

\textbf{Shared vs. separate positional embedding:} as can be seen in \Cref{tab:ablation_pe}, having two separate positional embeddings for the text and pose instead of having a shared one has a large effect on the results.

\textbf{Varying hidden and feedforward dimensions:} although the hidden and feedforward dimensions do not have a large effect, they still yield slightly better results (\Cref{tab:ablation_hidden,tab:ablation_ff}).

\begin{table}
  \centering
  \begin{tabular}{@{}c@{\hskip 0.5em}c@{\hskip 0.5em}c@{\hskip 0.5em}c@{\hskip 0.5em}c}
    \toprule
    Reference & \#steps & Rank 1 $\uparrow$ & Rank 5 $\uparrow$ & Rank 10 $\uparrow$ \\
    \midrule
    \multirow{3}{1.7cm}{\centering Prediction} & 5 & 0.09 & 0.16 & 0.28 \\
    & 10 & 0.08 & 0.2 & 0.35 \\
    & 20 & 0.09 & 0.22 & 0.35 \\
    \midrule
    \multirow{3}{1.7cm}{\centering Ground Truth} & 5 & 0.25 & 0.44 & 0.54 \\
    & 10 & 0.21 & 0.44 & 0.56 \\
    & 20 & 0.27 & 0.44 & 0.54 \\
    \bottomrule
  \end{tabular}
  \captionsetup{justification=centering}
  \caption{Ablation: different step amounts. Top: distance to prediction. Bottom: distance to ground truth pose.}
  \label{tab:dist2both_step}
\end{table}
\begin{table}
  \centering
  \begin{tabular}{@{}c@{\hskip 0.5em}c@{\hskip 0.5em}c@{\hskip 0.5em}c@{\hskip 0.5em}c}
    \toprule
    Reference & PE & Rank 1 $\uparrow$ & Rank 5 $\uparrow$ & Rank 10 $\uparrow$ \\
    \midrule
    \multirow{2}{1.7cm}{\centering Prediction} & Shared & 0.06 & 0.19 & 0.28 \\
    & Separate & 0.08 & 0.2 & 0.35 \\
    \midrule
    \multirow{2}{1.7cm}{\centering Ground Truth} & Shared & 0.16 & 0.37 & 0.51 \\
    & Separate & 0.21 & 0.44 & 0.56 \\
    \bottomrule
  \end{tabular}
  \captionsetup{justification=centering}
  \caption{Ablation: shared vs. separate positional embeddings. Top: distance to prediction. Bottom: distance to ground truth pose.}
  \label{tab:ablation_pe}
\end{table}

\begin{table}
  \centering
  \begin{tabular}{@{}c@{\hskip 0.5em}c@{\hskip 0.5em}c@{\hskip 0.5em}c@{\hskip 0.5em}c}
    \toprule
    Reference & Hidden dim & Rank 1 $\uparrow$ & Rank 5 $\uparrow$ & Rank 10 $\uparrow$ \\
    \midrule
    \multirow{3}{1.7cm}{\centering Prediction} & 64 & 0.06 & 0.19 & 0.35 \\
    & 128 & 0.08 & 0.2 & 0.35 \\
    & 256 & 0 & 0.002 & 0.007 \\
    \midrule
    \multirow{3}{1.7cm}{\centering Ground Truth} & 64 & 0.22 & 0.41 & 0.53 \\
    & 128 & 0.21 & 0.44 & 0.56 \\
    & 256 & 0.005 & 0.13 & 0.34 \\
    \bottomrule
  \end{tabular}
  \captionsetup{justification=centering}
  \caption{Ablation: different hidden dimensions. Top: distance to prediction. Bottom: distance to ground truth pose.}
  \label{tab:ablation_hidden}
\end{table}

\begin{table}
  \centering
  \begin{tabular}{@{}c@{\hskip 0.5em}c@{\hskip 0.5em}c@{\hskip 0.5em}c@{\hskip 0.5em}c}
    \toprule
    Reference & FF dim & Rank 1 $\uparrow$ & Rank 5 $\uparrow$ & Rank 10 $\uparrow$ \\
    \midrule
    \multirow{3}{1.7cm}{\centering Prediction} 
    & 512 & 0.08 & 0.23 & 0.36 \\
    & 1024 & 0.06 & 0.18 & 0.32 \\
    & 2048 & 0.08 & 0.2 & 0.35 \\    
    \midrule
    \multirow{3}{1.7cm}{\centering Ground Truth} 
    & 512 &  0.2 & 0.43 & 0.53 \\
    & 1024 & 0.19 & 0.42 & 0.54 \\
    & 2048 & 0.21 & 0.44 & 0.56 \\
    \bottomrule
  \end{tabular}
  \captionsetup{justification=centering}
  \caption{Ablation: different feedforward dimensions. Top: distance to prediction. Bottom: distance to ground truth pose.}
  \label{tab:ablation_ff}
\end{table}

\section{Loss function importance}
\begin{figure}[t]
    \centering
    \setlength{\tabcolsep}{0.5pt}
    \renewcommand{\arraystretch}{0.5}

    \begin{tabular}{ccc}

        original pose &
        MSE prediction &
        WMSE prediction \\
        
        \includegraphics[trim=0 0 8 0, clip, width=0.33\linewidth]{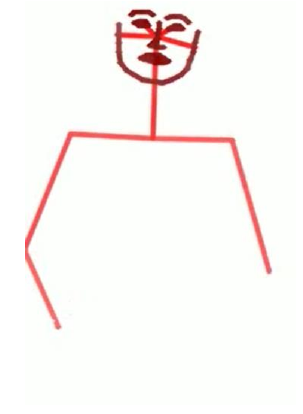} &
        
        \includegraphics[trim=4 0 0 0, clip, width=0.32\linewidth]{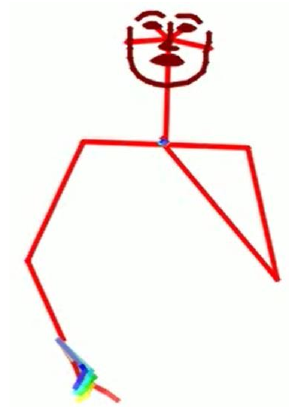} &

        \includegraphics[clip, width=0.3\linewidth]{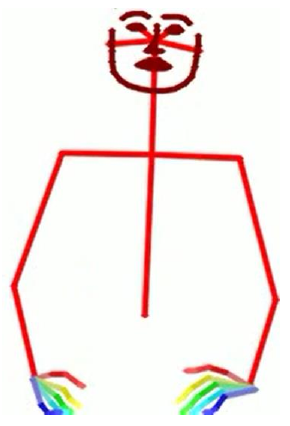} 
    \end{tabular}
    \caption{\textbf{MSE vs. weighted MSE loss example.} Left to right: original pose, pose generated by a model trained with MSE, pose generated by our model trained with weighted MSE. Regular MSE doesn't take missing keypoints into account, hence the model doesn't learn to generate a full pose.}
    \label{fig:mse_example}
    
\end{figure}

To show the importance of the weights in our weighted MSE loss function, we experiment with two other loss functions for the training of our model: regular MSE, and half-masked MSE---ignoring low confidence keypoints, but with equal weight for other keypoints. As our data contains many missing keypoints, a loss function that considers them is needed, so the model can learn to fill in the missing keypoints and predict a full pose. Therefore, when using regular MSE, as demonstrated in \Cref{fig:mse_example}, the model does not predict a full pose, and instead maps a lot of the keypoints to $(0,0)$.
The half-masked MSE loss performed better, but as keypoints with higher confidence are more likely to have correct locations, we wanted them to effect the loss of the model more than keypoints with low confidence, and indeed our masked MSE loss gave the best results. 

\section{Train and Inference Duration}
\label{supp:time}
\begin{table}
  \centering
  \begin{tabular}{ccc}
    \toprule
    \#steps & train (hrs) & inference (sec) \\
    \midrule
    5 & 9 & 0.03  \\
    10 & 20 & 0.06  \\
    20 & 39 & 0.12 \\
    \bottomrule
  \end{tabular}
  \captionsetup{justification=centering}
  \caption{Train and inference duration for different number of steps. Train time is in hours, inference is in seconds.}
  \label{tab:time_per_num_steps}
\end{table}
We train our models on one machine with 4 NVIDIA GeForce GTX TITAN X GPUs. The training and inference time for 2000 epochs over all languages is presented in \Cref{tab:time_per_num_steps} for different step amounts. As demonstrated in the table, doubling the number of steps doubles the train and inference duration as well. Having said that, the inference time for either number of steps is very low.

\section{Model Architecture Details}
\label{supp:arch}
\begin{figure*}
  \centering
   \includegraphics[width=\linewidth]{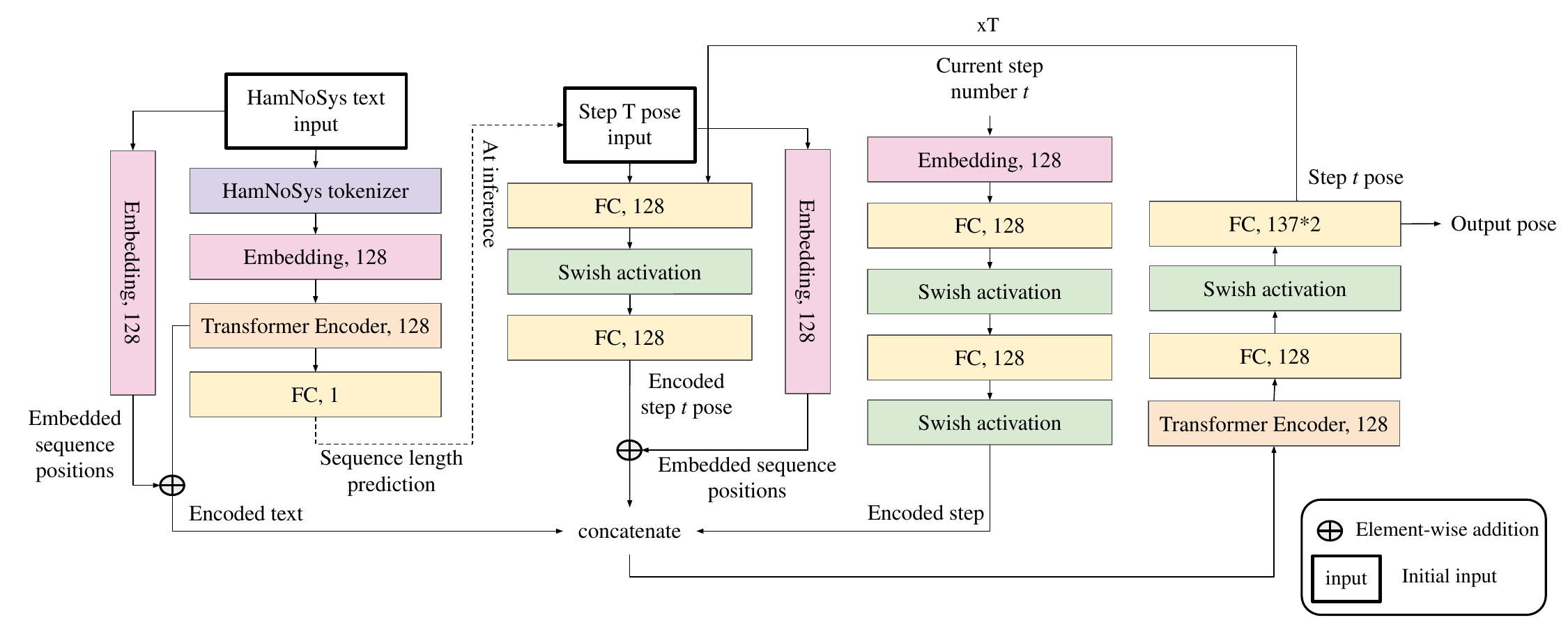}
   \caption{Detailed model architecture.}
   \label{fig:detailed_arch}
\end{figure*}

Our model is composed of two parts: the text processor and the pose generator. 
The text processor consists of:
\begin{itemize}
    \item HamNoSys tokenizer which converts each glyph into a unique identifier (token).
    \item Learned embedding layer with dimension 128 to embed the HamNoSys text.
    \item Learned embedding layer with dimension 128 to embed the text tokens positions (positional embedding).
    \item Transformer encoder \cite{vaswani2017attention} with 2 heads and depths 2, with a feedforward dimension of 2048.
    \item Fully connected layer which acts as the sequence length predictor, that gets the encoded text as input and returns a single predicted number.
\end{itemize}

The pose generator consists of:
\begin{itemize}
    \item Pose encoding: composed of two fully connected layers with dimension 128 for the hidden and output sizes, with a Swish \cite{ramachandran2017Swish} activation between them; and a positional embedding for the pose sequence locations, composed of a learned embedding layer. They are summed into pose embeddings of dimension 128.
    \item Step number encoding: composed of a learned embedding layer followed by two fully connected layers with Swish activations between them with hidden and output sizes of 128 as well.
    \item Encoding of the concatenation of all three encodings of the text, pose, and step. The encoding consists of a transformer encoder with 2 heads and depth 4, with a feedforward dimension of 2048, followed by a pose projection, which is composed of 2 fully connected layers with Swish activation between them, with hidden size of 128 and output size of the pose dimension, which is $137\times2$ in our case.
    
\end{itemize}

The pose is gradually generated over $T=10$ steps, where each step gets the output of the previous step as input. Finally, after T steps, the output of the model is the output of the pose projection. \\
A detailed overview of our model architecture is presented in \Cref{fig:detailed_arch}.

\section{Data}
\label{supp:data}
Our data consists of videos of Sign languages signs with their HamNoSys transcriptions. To use these videos as ground truth for pose sequence generation, we extract estimated pose keypoints from them using the OpenPose \cite{cao2017realtime} pose estimation model.
Each keypoint $k_i\in{K}$ consists of a 2D location $(x, y)$ and the confidence of the model in the location of that keypoint, $c_i$. The number of extracted keypoints is 137 per video frame, spanning the body ($K_B$, 25 keypoints), face ($K_F$, 70 keypoints), and hands ($K_H$, 21 keypoints per hand).

We process the keypoints further to use them, and define $c_{min}=0.2$ to be the minimal confidence for a keypoint to be considered as identified. As part of the pre-process, we define the following criteria:

\begin{equation}
\begin{multlined}
\quad \sum_{i\in{K_F}} c_i \leq c_{min}\cdot|K_F| \quad \textrm{or} \\ c_{r\_wrist}+c_{l\_wrist} \leq c_{min}
\end{multlined}
\label{eq:removal_criteria}
\end{equation}

and remove each leading or trailing frame for which they hold. Meaning, frames in which the face average keypoints confidence is less than $c_{min}$, or both hands are not identified, are removed.

\section{Loss Scaling Factor Derivation}
\label{supp:loss_deriv}
The pose generation process is gradual over T steps, where in each step we use a different step size as defined in \Cref{sec:method}. The step size at each time step depends on the chosen number of steps. Since we calculate a refinement loss for each step, $L_p$, to avoid affecting the learning rate when experimenting with different step values, we scale the loss by $ln(T)^2$. This scaling factor emerges from the following derivation:

\begin{equation} \label{step_size_derive}
\begin{aligned}
& \sum_{t=T-2}^0\alpha_t =  \sum_{t=T-2}^0\delta_t-\delta_{t+1} = \\ 
& \sum_{t=T-2}^0\log_T{(T-t)}-\log_T{(T-(t+1))} = \\
& \sum_{t=T-2}^0\log_T{\frac{T-t}{T-t-1}} = \\ 
& \sum_{t=1}^{T-1}\log_T{\frac{t+1}{t}} = \sum_{t=1}^{T-1}\frac{ln(\frac{t+1}{t})}{ln(T)} 
\end{aligned}
\end{equation}

We do not include $t = T-1$ in the sum since we define $\alpha_{T-1}$ to be the same constant regardless of the step size, to avoid illegal calculations.
As we can see from the equation above, in a full cycle of the pose generator, the denominator is the only part that depends on the step number, thus multiplying the loss by the square of the denominator eliminates the step number effect. 
Therefore, after scaling, the refinement loss term is: $ln(T)^2 L_p$.

\end{document}